\documentclass{article}

     \usepackage[preprint]{neurips_2025}

\usepackage[utf8]{inputenc} % allow utf-8 input
\usepackage[T1]{fontenc}    % use 8-bit T1 fonts
\usepackage{hyperref}       % hyperlinks
\usepackage{url}            % simple URL typesetting
\usepackage{booktabs}       % professional-quality tables
\usepackage{amsfonts, amsmath, bm}       % blackboard math symbols
\usepackage{nicefrac}       % compact symbols for 1/2, etc.
\usepackage{microtype}      % microtypography
\usepackage{xcolor}         % colors
\usepackage{tabularx}
\usepackage{placeins}

\title{Great GATsBi: Hybrid, Multimodal, Trajectory Forecasting for Bicycles using Anticipation Mechanism }

\author{
  Kevin Riehl\textsuperscript{1}, Shaimaa K. El-Baklish\textsuperscript{1}, Anastasios Kouvelas\textsuperscript{1}, Michail A. Makridis\textsuperscript{1} \\ % Coauthor \\
  \textsuperscript{1} Traffic Engineering Group, Institute for Transport Planning and Systems, ETH Zürich \\ % Affiliation \\
  Stefano-Franscini-Platz 5, Zurich, 8093, Switzerland \\ % Address \\
  kriehl@ethz.ch \\ % \texttt{email} \\
}

\begin{document}

\maketitle

%%%%%%%%%%%%%%%%%%%%%%%%%%%%%%%%%%%%%%%%%%%%%%%%%%%%%%%%%%%%
\begin{abstract}
    Accurate prediction of road user movement is increasingly required by many applications ranging from advanced driver assistance systems to autonomous driving, and especially crucial for road safety.
    Even though most traffic accident fatalities account to bicycles, they have received little attention, as previous work focused mainly on pedestrians and motorized vehicles.
    In this work, we present the \textit{Great GATsBi}, a domain-knowledge-based, hybrid, multimodal trajectory prediction framework for bicycles.
    The model incorporates both physics-based modeling (inspired by motorized vehicles) and social-based modeling (inspired by pedestrian movements) to explicitly account for the dual nature of bicycle movement.
    The social interactions are modeled with a graph attention network, and include decayed historical, but also anticipated, future trajectory data of a bicycles neighborhood, following recent insights from psychological and social studies.
    The results indicate that the proposed ensemble of physics models -- performing well in the short-term predictions -- and social models -- performing well in the long-term predictions -- exceeds state-of-the-art performance.
    We also conducted a controlled mass-cycling experiment to demonstrate the framework's performance when forecasting bicycle trajectories and modeling social interactions with road users. 
\end{abstract}

%%%%%%%%%%%%%%%%%%%%%%%%%%%%%%%%%%%%%%%%%%%%%%%%%%%%%%%%%%%%
\section{Introduction}

Accurate prediction of road user movement is crucial for various applications, including urban planning, infrastructure design, and intelligent transportation systems. 
Effective tracking and surveillance of road users can provide timely information, enabling smart signalized intersections~\cite{yu2023v2x}.
Moreover, precise forecasts of neighboring road users enhance safety systems in advanced driver assistance technologies, helping prevent collisions and accidents~\cite{kosaraju2019social}. 
Autonomous vehicles, in particular, frequently interact with neighboring road users and require accurate perceptions of both current and future positions to navigate safely~\cite{madjid2025trajectory}. 

Trends in urbanization, municipal sustainability efforts, increasing issues with congestion and emissions, and technological innovations drive the increasing use of active modes, like e-bikes and bicycles.
Each year, road traffic accidents result in approximately 1.24 million fatalities, with vulnerable road users (VRUs) and especially cyclists comprising the largest proportion of these deaths~\cite{toroyan2013launches,gao2021trajectory,zernetsch2016trajectory,bessi2023trajectory}.
Despite the increasing trend in severe traffic accidents involving bicycles and VRUs~\cite{de2020injuries,juhra2012bicycle}, their trajectory forecasting remains heavily underrepresented in the literature, that primarily focuses on pedestrians and motorized vehicles~\cite{huang2022survey,ding2023incorporating,schuetz2023review,madjid2025trajectory}.

Bicycle trajectory prediction is particularly challenging due to their hybrid behavioral characteristics. 
Bicycles and motorized vehicles alike, are subject to non-holonomic kinematic constraints and can reach relatively high speeds. 
Similarly, maneuver-related planning patterns such as following, lane changing and overtaking can be observed~\cite{gao2021trajectory}. 
However, unlike cars, bicycles are often less constrained by lane boundaries and can exhibit more flexible and sudden-change behaviors (in direction), similar to pedestrians~\cite{arlauskas2025,li2023two,brunner2024microscopic}. 
This combination of vehicle-like dynamics and pedestrian-like maneuverability complicates accurate trajectory prediction, especially in mixed-traffic environments~\cite{li2023two}.

Previous works on bicycle trajectory forecasting primarily modeled bicycle behavior similar to pedestrians.
By focusing on intent~\cite{pool2019context, gao2021trajectory}, sudden change behavior~\cite{li2023two} and multiple interactions with the neighborhood~\cite{huang2021lstm,li2023two} when modeling social context features, they neglect physical, kinematics-related properties of bicycles.
Moreover, previous efforts are characterized by data-driven approaches only, overlooking the potential of incorporating domain-knowledge with learning, and impeding interpretability~\cite{borghesi2020improving,dash2022review}.

To address the challenges of bicycle trajectory forecasting and related limitations, we propose the great \textit{GATsBi}, a hybrid, domain knowledge-based framework, that leverages both social and physical modeling, for multimodal trajectory forecasting of bicycles.
First, we model the social context as a graph structure using graph attention networks (GATs) and explicitly incorporate anticipation and perception decay following recent insights from behavioral modeling~\cite{hastie2022schematic,tarder2024brain,kinsky2020trajectory,tanke2023social,tanke2023social,ruan2024learning}.
Second, we combine kinematic and extended Kalman-filtering forecasting models to a deep ensemble embedding for the physics context.
Third, we follow a hybrid approach, that combines these two, domain-knowledge-based contextual features, to gain improvements in forecasting and enabling insights for interpretability.
Doing so, we account for the dual nature of bicycles, that share behavioral aspects with pedestrians and motorized vehicles.

We conducted a real-world, closed-loop, mass cycling experiment that enabled the observation of various interactions and bicycle maneuvers in varying traffic density contexts without the interference of road-contextual properties.
A comparative benchmark on the trajectory dataset supports the contributions of the proposed method when compared with common forecasting methods for pedestrians and vehicles.
The trajectory dataset, and implementation will be made publicly accessible in an open-source repository upon publication.

\begin{figure} [!ht]
    \centering
    \includegraphics[width=1.0\linewidth]{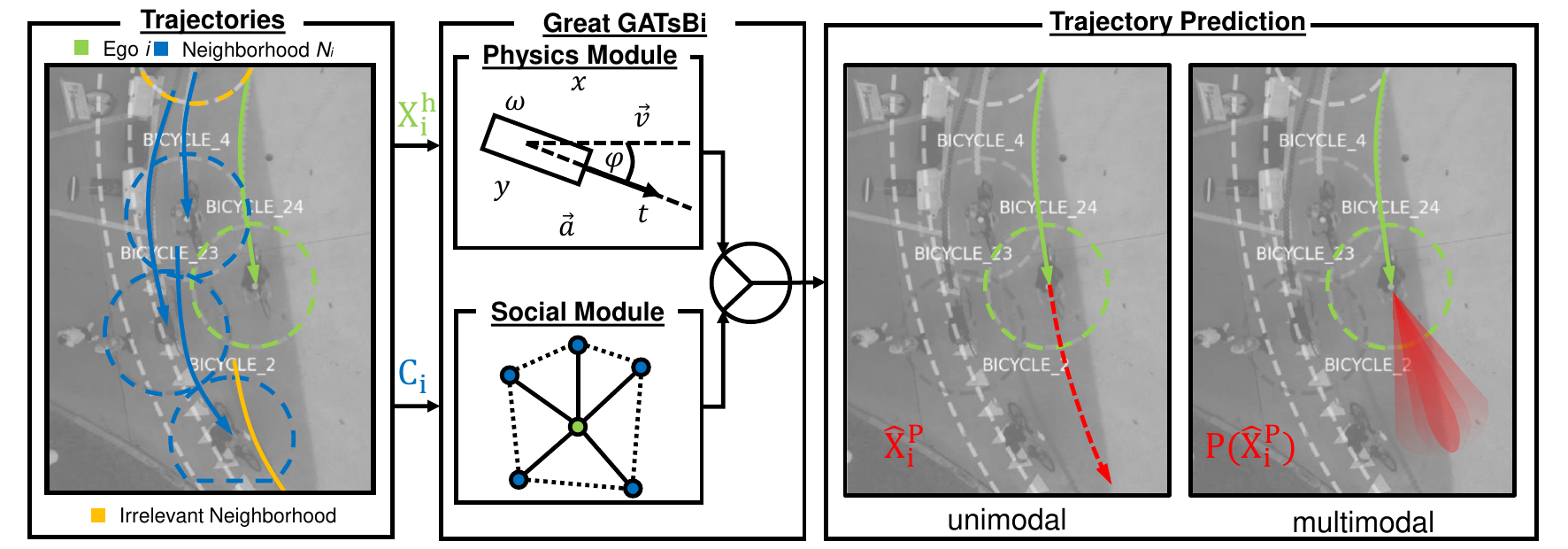}
    \caption{\textbf{Problem Statement: Trajectory Prediction.} This work is concerned with forecasting the trajectories of moving road users given their neighbors and their own (ego) historical trajectory information.}
    \label{fig:fig_1l}
\end{figure}

%%%%%%%%%%%%%%%%%%%%%%%%%%%%%%%%%%%%%%%%%%%%%%%%%%%%%%%%%%%%
\section{Related Work}

Trajectory forecasting typically involves contextual factors related to physics, road conditions, and social interactions. 
Prediction models can be categorized into physics-based methods, such a Kalman filtering approaches, and machine learning-based methods, which are better suited for capturing road and social context. 
These models can predict various types of outcomes, including a single trajectory (unimodal), a probabilistic distribution of possible future trajectories (multimodal), and driver intentions (maneuvers)~\cite{huang2022survey}.
Recent works especially leveraged the potential of machine and deep learning (ML) based approaches~\cite{ding2023incorporating}.

\textbf{Bicylce Trajectory Forecasting.}
Contrary to previous work that relied on kinematic modeling approaches, ~\cite{zernetsch2016trajectory} combined a simple physics model with a multi-layer perceptron (MLP) to achieve better predictions.
\cite{pool2017using} showed that combining road topology features with the bicycle's historical trajectory further enhances forecasts.
\cite{pool2019context, gao2021trajectory} highlighted the importance of learning the latent intention of the cyclist, which allows the model to leverage behavioral cues in trajectory prediction.
\cite{li2023two} expanded on this by differentiating between intentions and latent sudden change behavior to better account for behavioral heterogeneity.
%and proposed a two step approach to (i) learn a set of plausible trajectories and to (ii) select one of those.
\cite{huang2021lstm} demonstrated that incorporating neighborhood information to model multiple interactions between the cyclist and other road users can yield further improvements.
Besides, a growing branch of work explored fast-inference, learning-based sensor fusion of onboard sensors for equipped, smart bicycles~\cite{koornstra2023predicting,arlauskas2025,sass2023autonomous,bessi2023trajectory}.
Despite these advances, the former works are limited in three aspects. First, they neglected the multimodal nature of bicycle behavior which was shown to significantly improve forecasting accuracy in a pedestrian context~\cite{kosaraju2019social}.
Second, they focused on social context and overlooked the importance of systematic combination of physics-knowledge-based models.
Third, their neighborhood was modeled over-simplistically in the form of focal attention mechanisms.
To address these gaps, we propose a multimodal forecasting model based on Gaussian Mixture Models (GMM), explicitly combining physics-knowledge and social-knowledge to improve forecasting accuracy, and model the bicycle's social context as a graph using graph attention networks (GATs). 

\textbf{Human Motion Modeling.}
Traditionally, human motion modeling was often characterized by analytical modeling approaches.
Notably, \textit{Helbing's Social Force Model}~\cite{helbing1995social} explicitly takes surrounding individuals and human-human interactions into account when modeling human motion. 
Modern approaches yield significantly improved forecasts leveraging ML approaches, such as \textit{SocialLSTM}~\cite{alahi2016social}.
\textit{Social-BiGAT}~\cite{kosaraju2019social} highlighted the importance of multimodality in the context of motion modeling, and explicitly reflected the social context in a graph neural network (GNN).
In our model, we draw upon these advancements from pedestrian modeling and transfer them into the context of bicycles.

\textbf{Anticipation-Based Behavior Modeling.}
Recent brain studies indicate human perception systems anticipate the actions of others.
Consequently, studies from social sciences~\cite{ng2022learning} predict human motions based on the surrounding actors’ behaviors and interactional communication in dyadic conversations.
~\cite{tanke2023social} develops a \textit{Social Diffusion Model} for motion forecasting, anticipating social signals based on contextual human poses.
~\cite{ruan2024learning} presents a cooperative approach, where autonomous cars share local information for enhancing prediction accuracy, indicating the importance of shared perception in trajectory forecasting.
Moreover, findings indicate that hippocampus-related memory systems are subject to decay~\cite{hastie2022schematic,tarder2024brain,kinsky2020trajectory}.
Due to these findings and demonstrated advancements of behavioral anticipation in different domains, we explicitly include anticipation and perception decay to enhance the social context of trajectory forecasting beyond mere graph features, and thus to better extract social signals from the neighborhood.

\textbf{Knowledge-Based Machine Learning.}
A growing branch of literature is concerned with incorporating domain knowledge into data-driven ML approaches, yielding better forecasts as ensemble and enhancing interpretability~\cite{borghesi2020improving,dash2022review}.
In the context of trajectory prediction, social domain knowledge has been shown to yield improvements for pedestrians~\cite{korbmacher2022review,huang2022survey}, and physics domain knowledge has been shown to yield improvements for motorized road vehicles~\cite{liao2024physics,geng2023physics}, and drones~\cite{shukla2024trajectory,bianchi2024physics} recently.
Based on these insights, we aim to follow a hybrid approach and to combine social and physics domain-knowledge from pedestrians and vehicles to enhance bicycle trajectory forecasts.

\section{Methods: The Great GATsBi}

\subsection{Problem Definition}

The problem of trajectory forecasting manifests in the prediction $\mathcal{F}$ of entity $i$'s (hereafter called the \textit{ego}) future movement $X_i^{p}$ through space, given its past trajectory $X_i^{h}$ and contextual information $\mathcal{C}_i$: $\hat{X}_i^{p} = \mathcal{F} \{ X_i^{h} , \mathcal{C}_i \}$. 
Following previous works on pedestrians, bicycles, and motorized vehicles, we consider a two dimensional space with a historical observation horizon $t_{obs}$ and a prediction horizon $t_{pred}$, resulting in a past trajectory $X_i^{h} = \{ (x_i^{t}, y_i^{t} ) \in \mathbb{R}^2 | 0 \leq t \leq t_{obs} \} $ and a future trajectory $ X_i^{p} = \{ (x_i^{t}, y_i^{t} ) \in \mathbb{R}^2 | t_{obs} \leq t \leq t_{obs}+t_{pred} \} $.
In this work, we consider social contextual information $C_i = C_i^S$ only, excluding road-contextual information $C_i^R$. The social contextual information consists of the past trajectory $X_j^{h}$ of the ego's neighborhood $\mathcal{C}_i^S = \bigcup_{j\in\mathcal{N}_i} X_j^h$ (neighboring road users of ego bicycle $i$).
In the unimodal trajectory forecasting context, the prediction represents one trajectory $\mathcal{F} \in \mathbb{R}^2$, while in this work we consider the multimodal trajectory forecast, which is a probabilistic distribution across possible future trajectories $\mathcal{F} \in P(\mathbb{R}^2)$.

Common evaluation measures in the domain of trajectory prediction, that we will use in this work, include the average displacement error 
$ADE = \frac{1}{t_{pred}} \sum_{t} \| \hat{X}_i^{p} - X_i^{p}\|_2$, and the final displacement error $FDE = \| \hat{X}_{i,t_f}^{p} - X_{i,t_f}^{p}\|_2$ at time $t_f=t_{obs}+t_{pred}$. 

\subsection{Model Architecture}
\begin{figure} [!h]
    \centering
    \includegraphics[width=1.0\linewidth]{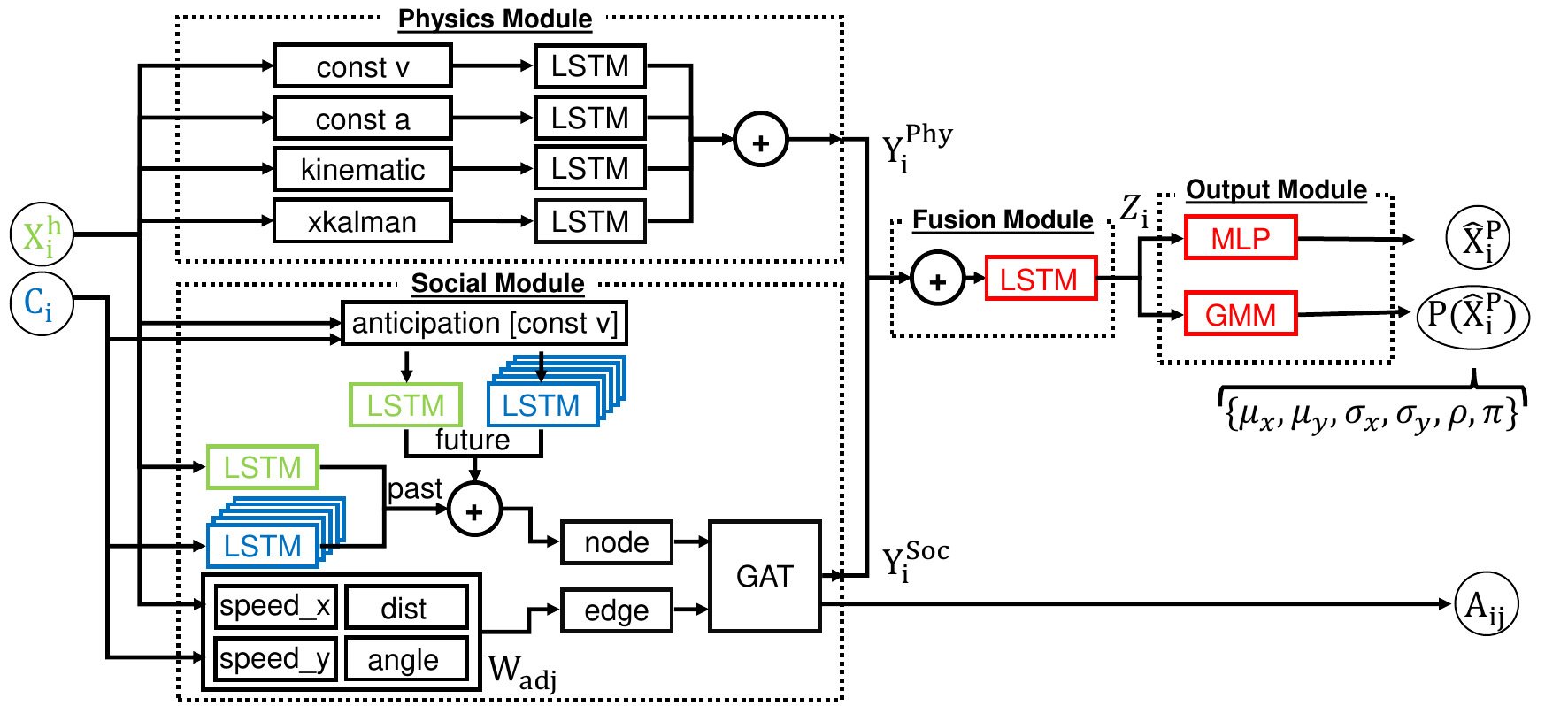}
    \caption{\textbf{Architecture for proposed Great GATsBi model.} The model consists of four modules. The physics module predicts the trajectory using common physical forecasting methods. The social module predicts the trajectory using a social graph structure, where the node features are a combination of past and anticipated future trajectories, and where the edge features are geometric adjacency features. The fusion module combines physics and social prediction. The output module transforms these forecasts back into the time domain, for unimodal or multimodal trajectory forecasting.}
    \label{fig:fig_2l}
\end{figure}

The proposed model comprises four main modules, as shown in Figure ~\ref{fig:fig_2l}. 
A \textit{physics module} utilizes domain knowledge of the vehicle-like bicycle dynamics to predict the possible physics-based future ego trajectory and encodes it into the latent space: $Y_i^{\text{Phy}}$. 
The \textit{social module} further uses the contextual information $C_i$ to encode the possible the ego's future trajectory into the same latent space: $Y_i^{\text{social}}$. 
The latent predictions, $Y_i^{\text{Phy}}$ and $Y_i^{\text{Soc}}$, are then amalgamated through the \textit{fusion module} and decoded to $Z_i$.
The \textit{output module} finally transforms the result either in the unimodal prediction $\hat{X}_{i}^{p}$, or into the multimodal prediction $P(\hat{X}_{i}^{p})$ in the form of Gaussian mixture models:$P(\hat{X}_{i}^{p}) = \{ \mu_x, \mu_y,\sigma_x, \sigma_y, \rho, \pi \}$. 
%The output space is the predicted ego trajectory $\hat{X}_i^P$ in the case of unimodal predictions. 
%The loss function in this case is the average displacement error; i.e $\mathcal{L}_{\text{unimodal}} = ADE$.
%Regarding multimodal predictions, the output space is the probabilistic distribution over all possible predicted ego trajectories $P(\hat{X}_i^P)$. 
%Here, we use a Gaussian mixture model of $K$ components; where each Gaussian component is characterized by the means $\mu_k = \begin{bmatrix} \mu_x^k & \mu_y^k \end{bmatrix}^T$, the covariance matrix $\Sigma_k = \begin{bmatrix} \sigma_{x, k}^2 & \rho_k \\ \rho_k & \sigma_{y, k}^2\end{bmatrix}$, and the mixture weight $\pi^k$. 
%The loss function used for this case is the negative log-likelihood loss.
%\begin{equation}
%    \mathcal{L}_{\text{GMM}} = - \mathbb{E}\log L(\hat{X}_i^P \mid X_i^P) = - \mathbb{E}\log \sum_{k=1}^K \pi_k \mathcal{N}(X_i^P \mid \mu_k, \Sigma_k)
%\end{equation}

\subsection{Physics Context Embedding}
In order to exploit the vehicle-like dynamics of the bicycle and existing domain knowledge, we combine the knowledge of four common physics-based models. 
\textit{const\_v} assumes a constant velocity point kinematic model for the bicycle, where the velocity is estimated from the last two frames of the historical trajectory. 
\textit{const\_a}, assumes a constant acceleration point kinematic model.
\textit{kinematic} assumes simple bicycle kinematics, characterized by position ($x,y$), direction ($\varphi$), speed ($v$):
%and angular velocity $\omega$: 
\begin{equation} \label{EQ:bicycle_kinematics_model}
    \dot{x} = v \cos\varphi, \qquad \dot{y} = v \sin\varphi, \qquad \dot{\varphi} = \frac{v}{L_B} \tan\delta
\end{equation}
where $\delta$ is the steering angle and $L_B$ is the length of the bicycle. 
\textit{xkalman} is an extended Kalman filtered~\cite{kalman1960new} prediction model based on the aforementioned kinematic model.
%, which uses the states $\bm{x}_{\text{xkalman}} = \begin{bmatrix} x & y & \varphi \end{bmatrix}^T$ and the same model as the bicycle kinematics model in \eqref{EQ:bicycle_kinematics_model} with the addition of $\dot{v} = a$; where $a$ is the linear acceleration input of the bicycle. 
All four physics-based forecasts for the ego trajectory are then encoded into latent space using separate long-short-term memory models (LSTMs). 
The encoded predictions are concatenated into the physical embeddings $Y_i^{\text{Phy}}$.
%: latent variable $Y_i^{\text{Phy}} = concat()$.
%\begin{equation}
%    Y_i^{\text{Phy}} = \bigcup_{w \in \mathcal{W}_{\text{Phy}}} LSTM_w (f_w(X_i^h))
%\mathbin\Vert for || 
%\end{equation}
%where $\mathcal{W}_{\text{Phy}}$ is the set of available physics-based models and $f_w(\cdot), \forall w \in \mathcal{W}_{\text{Phy}}$ is the corresponding prediction function.

\subsection{Social Context Embedding}
Social interactions are modeled in the \textit{social module}, following three assumptions based on recent findings in psychology and social sciences. 
First, human motions are affected by their neighborhood's social static context~\cite{huang2021lstm, li2023two,alahi2016social,kosaraju2019social}.
Second, humans anticipate the motion of their neighbors~\cite{ng2022learning,tanke2023social,ruan2024learning}. 
Third, humans capabilities to memorize past and to anticipate future trajectories are limited for longer time horizons~\cite{hastie2022schematic,tarder2024brain,kinsky2020trajectory}. 
To this end, we incorporate the following mechanisms when extracting contextual embeddings.

\textbf{Perception decay.} 
We give higher temporal ``attention'' to the sections of the past and anticipated trajectories closest to the current timestep. Therefore, we use exponential weights based on the temporal distance to the current timestep.
\begin{equation}
    D_h = \begin{bmatrix} e^{\lambda_h \cdot -t_{obs}} & \dots & e^{\lambda_h \cdot 0} \end{bmatrix}^T, \qquad D_p = \begin{bmatrix} e^{\lambda_p \cdot0} & \dots & e^{\lambda_p (t_{pred-1})} \end{bmatrix}^T
\end{equation}
where $\lambda_h \geq 0$ and $\lambda_p \leq 0$ are the decay parameters for the historical and predicted counterparts.

\textbf{Neighborhood anticipation.} 
We encode the neighborhood's historical trajectories $V^h_{\mathcal{N}_i}$ from $C_i^S$ using LSTMs to a latent space.
%\begin{equation}
%    V^h_i = LSTM_{\text{h,ego}}(X_i^h \cdot D_h), \qquad V^h_{\mathcal{N}_i} = \bigcup_{j \in \mathcal{N}_i} LSTM_{\text{h,Neigh}}(X_j^h \cdot D_h)
%\end{equation}
The neighborhood's future trajectories $V^p_{\mathcal{N}_i}$ are anticipated by the ego using \textit{const\_v} for modeling. 
These anticipated trajectories are transformed into the same latent space using.
%\begin{equation}
%    V^p_i = LSTM_{\text{p,ego}}(f_{\text{const\_v}}(X_i^h) \cdot D_p), \qquad V^h_{\mathcal{N}_i} = \bigcup_{j \in \mathcal{N}_i} LSTM_{\text{p,Neigh}}(f_{\text{const\_v}}(X_j^h) \cdot D_p)
%\end{equation}
\begin{equation}
    V^h_{\mathcal{N}_i} = \bigcup_{j \in \mathcal{N}_i} LSTM_{\text{h,}\mathcal{N}}(X_j^h \cdot D_h) \qquad 
    V^p_{\mathcal{N}_i} = \bigcup_{j \in \mathcal{N}_i} LSTM_{\text{p,}\mathcal{N}}(f_{\text{const\_v}}(X_j^h) \cdot D_p)
\end{equation}

\textbf{Social attention.} 
The historical and future trajectories of ego and neighborhood serve as node features of a social graph, that is modeled as a Graph Attention network (GAT)~\cite{kosaraju2019social}.
This enables the framework to enhance the ability to discern the relative importance of social interactions under the given static and anticipated social context. 
%\begin{subequations}
%\begin{align}
%    & V_i = V_i^h \mathbin\Vert V_i^p, \qquad V_j = V_j^h \mathbin\Vert V_j^p \; \forall j \in \mathcal{N}_i \\
%    & e_{ij} = a(W_{\text{adj}} V_i, W_{\text{adj}} V_j) \\
%    & \alpha_{ij} = \text{softmax}(e_{ij}) \\
%    & Y_i^{\text{Soc}} = \sum_{j \in \mathcal{N}_i} \alpha_{ij} W_{\text{adj}} V_j
%\end{align}
%\end{subequations}
%The node features are given by each cyclist's embedding for ego $V_i$ and neighbors $V_j \; \forall j \in \mathcal{N}_i$. 
The edge features are determined by an adjacency matrix $W_{\text{adj}}$, which includes relative distance, angle, and speeds between the ego and its neighborhood.
The resulting prediction of the GAT is denoted as the social embedding $Y_i^{\text{Soc}}$.
The attentions are returned in the weight matrix $A_{ij}$, and serve interpretability during inference and analysis.
%\textcolor{red}{The edge features are the graph adjacency matrix $W_{\text{adj}}$.} 
%The social attention weights are denoted by $\alpha_{ij}$ (i.e.\ for the interaction between ego cyclists $i$ and neighbor $j$) and the social context embeddings are $Y_i^{\text{Soc}}$ expressed in a latent space.
%The cyclist graph is maintained as fully connected; thereby enabling unrestricted interactions among all cyclists and avoiding any constraints on their ordering.

\subsection{Context Fusion \& Multimodal Decoder}
The \textit{fusion module} combines the physics and social context embeddings, $Y_i^{\text{Phy}}$ and $Y_i^{\text{Soc}}$, through concatenation and combined decoding through an LSTM to $Z_i$. 
The \textit{output module} projects $Z_i$ back into the original, Cartesian space (i.e., 2D $x-y$ space) using a Gauss Mixture Model layer for multimodal inference, and optionally a multilayer perceptron (MLP) for unimodal trajectory forecasts. 
For evaluation purposes in the multimodal forecast, we calculate the estimated trajectory $\hat{X}_i^P$ as the expected trajectory given the learned probability $P(\hat{X}_i^P)$ as follows:  
\begin{equation} \label{eq:expected_sampling}
    \hat{X}_i^P = \mathbb{E}_{\hat{X}} [ \hat{X}_i^P ] = \int \hat{X}_i^P P(\hat{X}_i^P) d\hat{X}_i^P
\end{equation}
%The \textit{fusion module} combines the physics and social context embeddings, $Y_i^{\text{Phy}}$ and $Y_i^{\text{Soc}}$, through concatenation and temporal repetitions. 
%Then, the combined context is decoded through an LSTM and fed into the \textit{output module}. The \textit{output module} projects the output back into the original space (i.e. 2D $x-y$ space). For unimodal predictions, the \textit{output module} is an MLP.
%\begin{equation}
%    \hat{X}_i^p = MLP_{\text{unimodal}}(LSTM(Y_i^{\text{Phy}} \mathbin\Vert Y_i^{\text{Soc}}))
%\end{equation}
%For multimodal predictions, the MLP expresses a $K$-component GMM instead.
%\begin{equation}
%    \{ \mu_1, \Sigma_1, \dots, \mu_K, \Sigma_K  \} = MLP_{\text{GMM}}(LSTM(Y_i^{\text{Phy}} \mathbin\Vert Y_i^{\text{Soc}}))
%\end{equation}

%%%%%%%%%%%%%%%%%%%%%%%%%%%%%%%%%%%%%%%%%%%%%%%%%%%%%%%%%%%%
\section{Experiments}

\subsection{Experiment Setup}

\textbf{Mass Cycling Experiment Dataset.}
% Present briefly dataset and mass cycling experiment.
We conducted a controlled, real-world, closed-loop, mass cycling experiment on a circular track.
The experiment included over more than 25 unique bicyclists, and was conducted at varying traffic density contexts (from 6 to 22 simultaneous bikes on the track) over one hour (in total more than 16,054 bicycle-frame observations).
Previous datasets had several limitations, as they were small in sample size~\cite{zernetsch2016trajectory}, or recorded at road intersections that complicate geometric behavior~\cite{gao2021trajectory,li2023two,huang2021lstm}.
\textit{Tsinghua-Daimler}~\cite{li2016new} or \textit{Eurocity Persons}~\cite{braun2019eurocity} offered recordings of low temporal resolution and unavailable coordinate transforms into Cartesian space.
\textit{Stanford Drone}~\cite{robicquet2016learning} is of high quality, but includes a complex scene with various obstacles on the university campus and mixed traffic scenarios.
Au contraire, our controlled experiment design allowed to exclude interference of road-contextual properties, and to study physical and social dynamics only. 
Therefore, \textit{Great GATsBi} was evaluated on our mass cycling trajectory dataset.
To analyse the contribution of the proposed anticipation mechanism on social embeddings in human motion predition, GATsBi was further evaluated on two pedestrian datasets (\textit{ETH} and \textit{HOTEL}~\cite{pellegrini2009you}).

% ~\cite{pool2019context}  EGO ONLY + INTENTION
%     - Stanford Drone (SD) dataset [9], 

% ~\cite{huang2021lstm}  NEIGHBORHOOD + DYNAMICS + ROAD GEOMETRY
%     - stanford drone dataset

\textbf{Implementation Details.}
% Talk about dimensions of models (hidden state, neural network neurons, etc)
% Talk about some more details of Neural Network.
For the experiments, we choose a historical horizon of 100 observations (equals 4 seconds) and different prediction horizons of $[ 25, 50, 75, 100]$ observations (equals 1,2,3,4 seconds).
At most five neighbors at a distance below 20m are considered.
All LSTM networks had a fixed hidden state dimension of 64, and the \textit{MLP} consisted of three sequential layers (linear + activation + linear) with a rectified linear (\textit{ReLU}, $\alpha=0.2$) unit as activation function. 
The \textit{GAT} network operates on a fully-connected graph~\cite{kosaraju2019social} using a single \textit{GAT} layer, uses a \textit{LeakyReLU} activation function for attention, and a dropout layer ($\phi=0.1$) for weight regularization.
% Talk about software
% Talk about optimizer used.
% Talk about hardware used and runtime.
% Talk about loss function for unimodal and multimodal.
% Talk about prediction and horizon length and timesteps in between.
\textit{Great GATsBi} was implemented in \textit{Python}(v3.11.6) using the \textit{Pytorch} deep learning library~\cite{paszke2019pytorch}. 
The network was trained using the \textit{Adam optimizer}~\cite{kingma2014adam} at a learning rate of $1 \times10^{-3}$, on a server with \textit{NVIDA RTX 4090} hardware and \textit{CUDA} (v11.6.0).
The first 50 epochs (calculated in less than an hour) were considered during training, and the models that performed best across all five train/test splits on average were selected for validation.
For unimodal forecasts, \textit{ADE} was used as loss function, for multimodal forecasts \textit{Gaussian Mixture Model Negative Log-Likelihood} was used as loss function.

\subsection{Main Results}
% Show results, compare approach with other methods

\begin{figure} [!ht]
    \centering
    \includegraphics[width=0.93\linewidth]{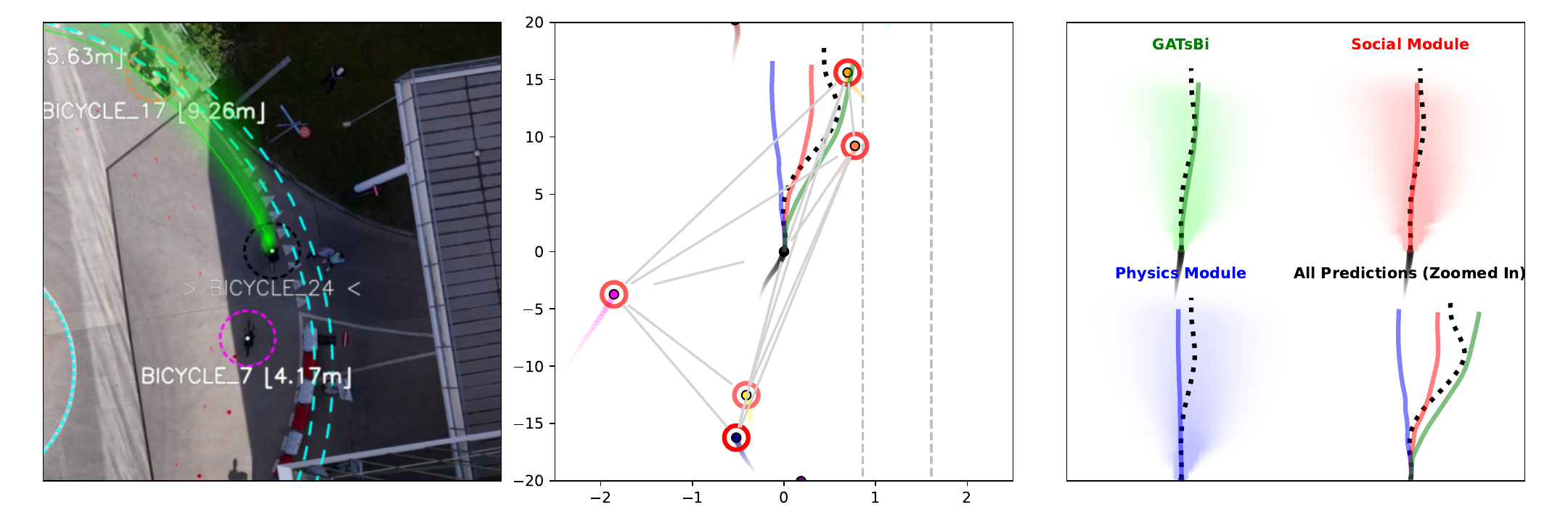}
    \caption{\textbf{Inference Result: Trajectory Prediction.} (left) annotated drone view; (middle) same scene in lane coordinates (relative to ego), ego's attention to neighbors represented by intensity of red circles, true trajectory as black dashed line, predicted trajectories in green (GATsBi's), red (social module), and blue (physics module); (right) probability distributions of multimodal forecasts.
    %multimodal forecast of different models show how GATsBi achieves more accurate prediction and lower variance.
    }
    \label{fig:fig_3l}
\end{figure}

\begin{figure} [!ht]
    \centering
    \includegraphics[width=0.93\linewidth]{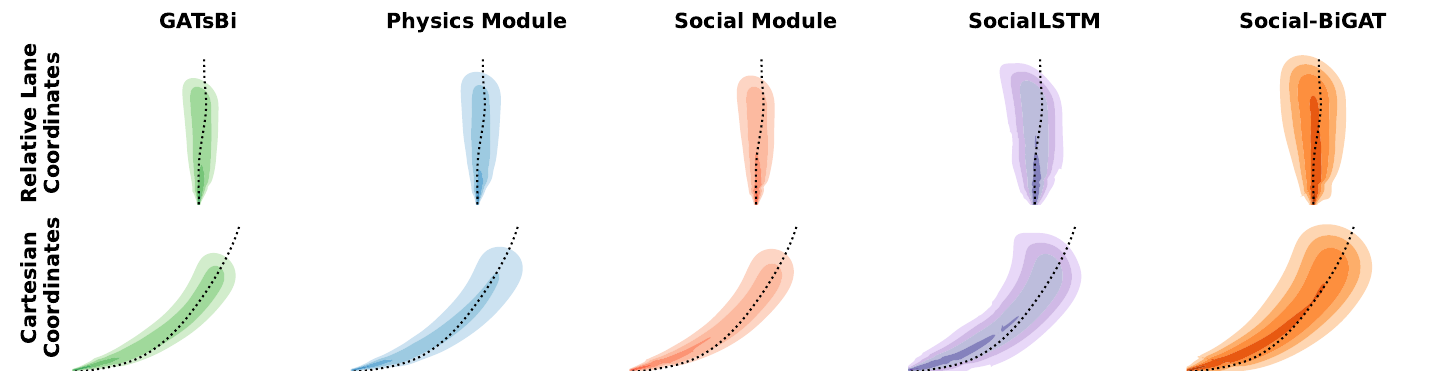}
    \caption{\textbf{Uncertainty Of Multimodal Forecasting Models.} Black dashed line represents true future trajectory. GATsBi and its components (social and physics module) both reduce uncertainty.}
    \label{fig:fig_uncertainty}
\end{figure}

Figure~\ref{fig:fig_3l} showcases the multimodal forecast improvements of \textit{GATsBi}.
While the physics module and social module alone are not able to capture the pending overtaking maneuver within the next four seconds sufficiently, the ensemble approach of both is.
The \textit{physics\_module} predicts a straight-ahead motion, which may be true for the short term. However, the long-term prediction is dominated by social interactions; where both \textit{GATsBi} and \textit{social\_module} can predict the curve maneuver that is brought about by the overtaking cyclist (i.e.\ bicycle 7 in the left of Figure~\ref{fig:fig_3l}).
Besides better forecasts, the density of the ensemble's probability distribution is higher, reducing uncertainty in forecasts. 
Figure~\ref{fig:fig_uncertainty} displays the probability distribution of the multimodal forecasts for the same scene.
\textit{GATsBi} and its sub modules achieve smaller uncertainty when compared with the baseline models \textit{SocialLSTM} and \textit{Social-BiGAT}. 

\FloatBarrier

Table~\ref{tab:main_results} compares \textit{GATsBi} with physics-based baseline models (\textit{const\_v}, \textit{const\_a}, \textit{kinematics}, and \textit{xkalman}) and social, learning-based baseline models that capture social interactions from pedestrian prediction literature (\textit{SocialLSTM} and \textit{Social-BiGAT}). 
The comparison reveals insights across different prediction horizons for ADE and FDE evaluation metrics.
%In addition, we compare both the \textit{physics\_module} and \textit{social\_module} of \textit{GATsBi} separately. 
%We report both ADE and FDE metrics on various prediction horizon lengths; i.e\ $t_{pred} \in \{ 1, 2, 3, 4\}$ seconds.
%
Despite their simplicity, the physics-based models achieve relatively accurate predictions on the short term, but worsen with longer horizons.
This is concordance to previous research~\cite{huang2022survey}.
Interestingly, the simplest of all, \textit{const\_v}, exhibits the lowest ADE and FDE over the 4 physics-based baselines for all horizons, as well as the lowest standard deviations. 
Coupled with its simplistic and interpretable nature, this motivates our choice for such a model to be used within our neighborhood anticipation contextual embeddings. 
As expected, the ensemble model \textit{physics\_module} outperform the physics-based baselines as it learns how to combine such predictions in the latent space of the encoder-decoder architecture used.
The more complex, learning- and social-based models achieve improvements in forecasting accuracy, especially at longer forecasting horizons, where again the ensemble model \textit{social\_module} achieves the best forecasts.
The performances of physics and social models reflect the dual nature of bicycle movement behavior.
\textit{SocialLSTM} performs better on the longer prediction horizons (3--4 seconds) than \textit{Social-BiGAT} on the FDE metric; while \textit{Social-BiGAT} is better for the shorter horizons for both ADE and FDE. 
The reason might be the different social graphs used by each model. 
\textit{Social-BiGAT} assumes a fully connected graph for the neighborhood, while \textit{SocialLSTM} allows only adjacent neighbors to interact through social pooling layer.
%; therefore, the input order of the cyclists may affect the results. 
Notably, the proposed \textit{social\_module} shows lower error metrics than both \textit{SocialLSTM} and \textit{Social-BiGAT}; highlighting the potential of included psychological model characteristics (i.e.\ anticipation and perception decay). 
Finally, \textit{GATsBi}, the combination of the \textit{physics\_module} and the \textit{social\_module}, significantly improves prediction accuracies, highlighting the potential of included psychological model characteristics (i.e. anticipation and perception decay). 
Notably, the combination of domain knowledge from social- and physics-context improve forecasts especially for longer time horizons (2s to 4s).

\begin{table}[!ht]
    \caption{\textbf{Forecasting Benchmark on Mass Cycling Dataset.} 
    Evaluation metrics (ADE and FDE) reported in average and standard deviation (in brackets) across all train/test splits for four different prediction horizons (1s to 4s). Bold numbers mark the best forecasting performance.}
    \label{tab:main_results}
    \centering
    \begin{tabularx}{\textwidth}{
        l
        c
        >{\centering\arraybackslash}X
        >{\centering\arraybackslash}X
        >{\centering\arraybackslash}X
        >{\centering\arraybackslash}X
        c
        >{\centering\arraybackslash}X
        >{\centering\arraybackslash}X
        >{\centering\arraybackslash}X
        >{\centering\arraybackslash}X
    }
        \toprule
        \textbf{Method} & & \multicolumn{4}{c}{\textbf{ADE}} & & \multicolumn{4}{c}{\textbf{FDE}} \\
        & & 1s & 2s & 3s & 4s & & 1s & 2s & 3s & 4s \\
        \cmidrule(lr){3-6} \cmidrule(lr){8-11}
        \textbf{Physics} & & & & & & & & & & \\
        \;\; const\_v & & 0.1080 \newline [0.0076] & 0.2818 \newline [0.0194] & 0.5460 \newline [0.0444] & 0.9406 \newline [0.1059] & & 0.2592 \newline [0.0182] & 0.6568 \newline [0.0436] & 1.5245 \newline [0.1787] & 2.7275 \newline [0.4278] \\
        \;\; const\_a & & 0.1281 \newline [0.0118] & 0.5504 \newline [0.0482] & 1.2951 \newline [0.1180] & 2.3929 \newline [0.2292] & & 0.3934 \newline [0.0346] & 1.6373 \newline [0.1422] & 4.0117 \newline [0.3857] & 7.3837 \newline [0.7451] \\
        \;\; kinematics & & 0.1103 \newline [0.0088] & 0.3942 \newline [0.0364] & 0.8914 \newline [0.0905] & 1.6309 \newline [0.1795] & & 0.3027 \newline [0.0260] & 1.1047 \newline [0.1068] & 2.7238 \newline [0.3056] & 4.9800 \newline [0.5935] \\
        \;\; xkalman & & 0.1445 \newline [0.0122] & 0.3269 \newline [0.0242] & 0.5967 \newline [0.0512] & 0.9948 \newline [0.1146] & & 0.3068 \newline [0.0235] & 0.7154 \newline [0.0492] & 1.5887 \newline [0.1904] & 2.7913 \newline [0.4417] \\
        \;\; $\ast$ \textnormal{physics\_module} & & 0.0802 \newline [0.0057] & 0.2263 \newline [0.0140] & 0.4513 \newline [0.0365] & 0.8045 \newline [0.0924] & & 0.2110 \newline [0.0136] & 0.5335 \newline [0.0313] & 1.3292 \newline [0.1714] & 2.4936 \newline [0.3703] \\
        \\
        \textbf{Social} & & & & & & & & & & \\
        \;\; SocialLSTM & & 0.0876 \newline [0.0071] & 0.2487 \newline [0.0133] & 0.4762 \newline [0.0359] & 0.8214 \newline [0.0911] & & 0.2141 \newline [0.0162] & 0.5479 \newline [0.0332] & 1.2829 \newline [0.1674] & 2.3770 \newline [0.4008] \\
        \;\; Social-BiGAT & & 0.0702 \newline [0.0068] & 0.2240 \newline [0.0139] & 0.4586 \newline [0.0377] & 0.8069 \newline [0.0898] & & 0.1914 \newline [0.0138] & 0.5242 \newline [0.0304] & 1.3234 \newline [0.1302] & 2.5356 \newline [0.3435] \\
        \;\; $\ast$ \textnormal{social\_module} & & \textbf{0.0629} \newline [0.0057] & 0.2101 \newline [0.0137] & 0.4284 \newline [0.0343] & 0.7834 \newline [0.0838] & & \textbf{0.1749} \newline [0.0838] & 0.4941 \newline [0.0309] & 1.2761 \newline [0.1376] & 2.4732 \newline [0.2935] \\
        \\
        $\ast$ \textbf{Great GATsBi} & & 0.0715 \newline [0.0066] & \textbf{0.2078} \newline [0.0130] & \textbf{0.4181} \newline [0.0354] & \textbf{0.7543} \newline [0.0960] & & 0.1893 \newline [0.0153] & \textbf{0.4891} \newline [0.0258] & \textbf{1.2641} \newline [0.1762] & \textbf{2.3827} \newline [0.4103] \\
        \bottomrule
    \end{tabularx}
\end{table}

\begin{figure} [!ht]
    \centering
    \includegraphics[width=1.0\linewidth]{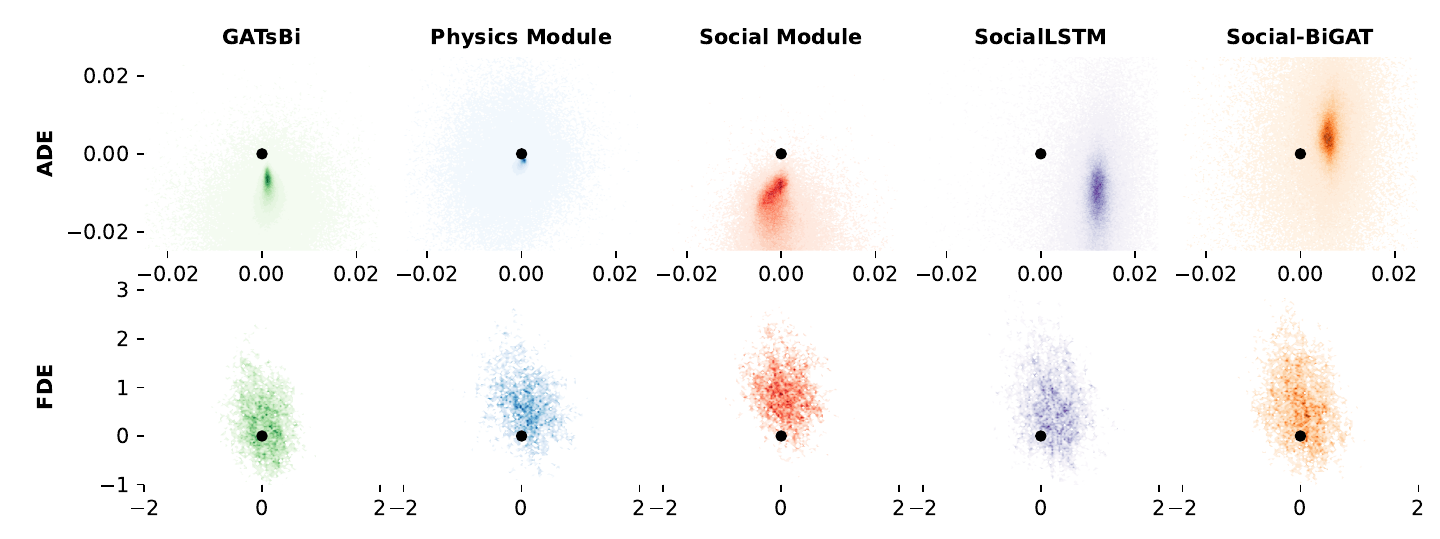}
    \caption{\textbf{ADE and FDE Error Distributions.} Two-dimensional distribution of ADE and FDE errors across all frame-bicycle combinations for 100s forecasts. Errors displayed in meters. The examined models differ in their bias, variance, and error distribution. Black point represents no error case.}
    \label{fig:fig_error}
\end{figure}

\FloatBarrier

Figure~\ref{fig:fig_error} analyzes the prediction error distribution (both in ADE and FDE terms).
In terms of average errors (ADE), one can observe biases in the forecasts, for all models except for the physics module. 
Especially \textit{SocialLSTM} and \textit{Social-BiGAT} obtain large biases in horizontal terms.
In terms of final errors (FDE), all models are bias free but differ in their variance, again \textit{GATsBi} yields the lowest uncertainty and highest accuracy.

\FloatBarrier
\subsection{Ablation Study}
% What happens when we remove certain features

To better understand the contributions of the elements of the proposed \textit{GATsBi} and anticipation mechanism, several ablation studies were conducted, as shown in Table~\ref{tab:ablation}.

The ablation of the multimodality in the output module generally worsens the forecasts. 
Multimodal models capture the multimodality and unpredictability of human future behavior, and therefore yield better results, as shown by previous studies~\cite{kosaraju2019social}. 
The ablation of the anticipation highlights the importance of this mechanism as part of our framework, especially for forecasts of longer time horizons.
The ablation of decays achieves worse outcomes as well.
Finally, the ablation of the full-connectedness of the social graph (down to a star-connected graph for the adjacency matrix) worsens predictions for long horizons.

To summarize, the ablations justify the design choices of the anticipation mechanism, and supports previous findings on human perception processes and motion forecasting from the psychological and social science literature.

\begin{table}[!ht]
    \caption{\textbf{Great GATsBi Ablations on Mass Cycling Dataset.} 
    Evaluation metrics (ADE and FDE) reported in average and standard deviation (in brackets) across all train/test splits for four different prediction horizons (1s to 4s).}
    \label{tab:ablation}
    \centering
    \begin{tabularx}{\textwidth}{
        l
        c
        >{\centering\arraybackslash}X
        >{\centering\arraybackslash}X
        >{\centering\arraybackslash}X
        >{\centering\arraybackslash}X
        c
        >{\centering\arraybackslash}X
        >{\centering\arraybackslash}X
        >{\centering\arraybackslash}X
        >{\centering\arraybackslash}X
    }
        \toprule
        \textbf{Ablations} & & \multicolumn{4}{c}{\textbf{ADE}} & & \multicolumn{4}{c}{\textbf{FDE}} \\
        & & 1s & 2s & 3s & 4s & & 1s & 2s & 3s & 4s \\
        \cmidrule(lr){3-6} \cmidrule(lr){8-11}
        % \textbf{With Physics Module} & & & & & & & & & & \\
        \;\; unimodal & & 0.0757 \newline [0.0059] & 0.2180 \newline [0.0125] & 0.4302 \newline [0.0343] & 0.7760 \newline [0.0945] & & 0.1948 \newline [0.0138] & 0.5051 \newline [0.0253] & 1.2286 \newline [0.1554] & 2.3142 \newline [0.4071] \\	
        \;\; no anticipation & & 0.0692 \newline [0.0063] & 0.2099 \newline [0.0121] & 0.4267 \newline [0.0257] & 0.7735 \newline [0.0795] & & 0.1868 \newline [0.0141] & 0.4924 \newline [0.0248] & 1.2962 \newline [0.1117] & 2.4673 \newline [0.3420] \\        
        \;\; no decay & & 0.0707 \newline [0.0063] & 0.2074 \newline [0.0125] & 0.4204 \newline [0.0338] & 0.7727 \newline [0.0966] & & 0.1893 \newline [0.0130] & 0.4901 \newline [0.0260] & 1.2824 \newline [0.1488] & 2.5016 \newline [0.4263] \\
        \;\; star-connected & & 0.0690 \newline [0.0086] & 0.2078 \newline [0.0127] & 0.4347 \newline [0.0368] & 0.7880 \newline [0.0810] & & 0.1873 \newline [0.0175] & 0.5006 \newline [0.0324] & 1.3159 \newline [0.1234] & 2.5150 \newline [0.2881] \\
        \\
        \textbf{Great GATsBi} & & 0.0715 \newline [0.0066] & 0.2078 \newline [0.0130] & 0.4181 \newline [0.0354] & 0.7543 \newline [0.0960] & & 0.1893 \newline [0.0153] & 0.4891 \newline [0.0258] & 1.2641 \newline [0.1762] & 2.3827 \newline [0.4103] \\
        % \\
        % \textbf{Without Physics Module} & & & & & & & & & & \\
        % \;\; no anticipation & & 0.0661 \newline [0.0087] & 0.2030 \newline [0.0164] & 0.4217 \newline [0.0301] & 0.7802 \newline [0.1037] & & 0.1795 \newline [0.0185] & 0.4879 \newline [0.0347] & 1.2807 \newline [0.1392] & 2.4736 \newline [0.4512] \\        
        % \;\; star-connected & & 0.0689 \newline [0.0077] & 0.2079 \newline [0.0253] & 0.4292 \newline [0.0417] & 0.7870 \newline [0.0949] & & 0.1867 \newline [0.0175] & 0.4957 \newline [0.0482] & 1.2969 \newline [0.1570] & 2.4383 \newline [0.3950] \\	
        % \;\; no decay & & 0.0644 \newline [0.0064] & 0.2125 \newline [0.0145] & 0.4325 \newline [0.0334] & 0.7892 \newline [0.0856] & & 0.1773 \newline [0.0149] & 0.4922 \newline [0.0318] & 1.3014 \newline [0.1318] & 2.4954 \newline [0.3966] \\						
        % \;\; unimodal & & 0.0713 \newline [0.0052] & 0.2255 \newline [0.0134] & 0.4400 \newline [0.0312] & 0.7794 \newline [0.0994] & & 0.1871 \newline [0.0127] & 0.5060 \newline [0.0213] & 1.2391 \newline [0.1564] & 2.3037 \newline [0.4107] \\
        \bottomrule
    \end{tabularx}
\end{table}

\subsection{Generalization Capability for Pedestrians}
% Experiment on Pedestrian Dataset, Vehicle Dataset

Table~\ref{table_3} evaluates the pedestrian forecasting accuracy of the proposed \textit{social\_module} and the related anticipation mechanism.
Two common pedestrian datasets (\textit{ETH} and \textit{HOTEL}~\cite{pellegrini2009you}) serve for this purpose.
The results show, that the simplest model, \textit{SocialLSTM}, outperforms all other models, including \textit{Social-BiGAT}.
One possible explanation is the relatively small size of these pedestrian datasets, compared to the relatively large complexity of \textit{Social-BiGAT} and the proposed models.
Yet, the proposed ensemble \textit{GATsBi} achieves second best results for all prediction horizons, not far away from those of \textit{SocialLSTM}.
%A comparison of the variances shows that \textit{GATsBi} can reduce forecast uncertainty.
Even though one might assume that social contexts primarily determine pedestrian motion, the \textit{physics\_module} performs better than the \textit{social\_module} in the ETH dataset.
The overall robust performance of \textit{GATsBi} and uncertainty reduction indicate the potential of the proposed anticipation mechanism for human motion forecasting, not only for bicycles but also for pedestrians.

\begin{table}[!ht]
    \caption{\textbf{Forecasting Benchmark on Pedestrian Datasets.} 
    Evaluation metric ADE reported for two benchmark datasets and for four forecasting horizons (0.8s to 4s).}
    \label{table_3}
    \centering
    \begin{tabularx}{\textwidth}{
        l
        c
        >{\centering\arraybackslash}X
        >{\centering\arraybackslash}X
        >{\centering\arraybackslash}X
        >{\centering\arraybackslash}X
        c
        >{\centering\arraybackslash}X
        >{\centering\arraybackslash}X
        >{\centering\arraybackslash}X
        >{\centering\arraybackslash}X
    }
        \toprule
        \textbf{Method} & & \multicolumn{4}{c}{\textbf{ADE (ETH)}} & & \multicolumn{4}{c}{\textbf{ADE (HOTEL)}} \\
        & & 0.8s & 1.6s & 2.4s & 4.0s & & 0.8s & 1.6s & 2.4s & 4.0s \\
        \cmidrule(lr){3-6} \cmidrule(lr){8-11}

        \;\; SocialLSTM & & 0.0150  & 0.0249  & 0.0372  & 0.0744  & & 0.0375 & 0.0621 & 0.0901 & 0.1788  \\
        \;\; Social-BiGAT & & 0.0541  & 0.0921 & 0.1384 & 0.2518 & & 0.0437 & 0.0787 & 0.1083 & 0.2130 \\
        \;\; $\ast$ \textnormal{physics\_module} & & 0.0290 & 0.0386 & 0.0417 & 0.0814  & & 0.0483 & 0.0760 & 0.1105 & 0.2150  \\
        \;\; $\ast$ \textnormal{social\_module} & & 0.0545  & 0.0970  & 0.1063  & 0.2319  & & 0.0432 & 0.0763 & 0.1095 & 0.2117  \\
        \;\; $\ast$ \textbf{Great GATsBi} & & 0.0420 & 0.0330 & 0.0479 & 0.0800  & & 0.0459 & 0.0737 & 0.1135 & 0.2170  \\
        \bottomrule
    \end{tabularx}
\end{table}

\section{Conclusion}
In this work, we proposed \textit{GATsBi} for multimodal trajectory prediction of bicycles, which despite their high fatality rates in road accidents experienced little attention in the literature. 
Contrary to previous approaches, \textit{GATsBi} amalgamates domain-knowledge from social interactions and physical kinematics in a systematic, deep-ensemble way.
Inspired by recent insights from the social science and psychology literature, we propose an anticipation mechanism to complement existing, graph-modeling of the social context.
This anticipation mechanism reflects the anticipating nature of humans during decision making, and their limited, decaying perception at longer time horizons.
Through evaluations on the conducted, controlled mass cycling experiment, we demonstrated the ability to generalize the complex, dual nature of bicycle motion, that is characterized by both similarities with pedestrians and cars.
At the same time, the proposed method performs robustly well for pedestrians.
Furthermore, ablation studies highlighted the contributions of the novel anticipation mechanism in the proposed architecture.
As such \textit{GATsBi} achieves lower bias and variance for more realistic bicycle trajectory forecasts, making it an invaluable component to safety-critical autonomous driving applications.

Future works might explore mixed traffic scenarios into more detail, such as providing neighbor classes to the social context, in order to enhance predictions of both pedestrians and bicycles interacting with each other.
As most fatal accidents with bicycles happen at road intersections and with larger vehicles such as trucks, further studies should include road geometries and better sensing for perception, as well as trajectory-forecasting-informed risk modeling.

%features for social interactions via GATs and a deep ensemble of physics-based models. 
%It also includes psychological modeling components for neighborhood anticipation and perception decay. 
%By utilizing data from our conducted mass-cycling experiment, we show that \textit{GATsBi} can improve prediction accuracy over various prediction horizon lengths and reduce variance (i.e.\ uncertainty of prediction). 
%Furthermore, ablation studies highlighted the contributions of the various modeling components of \textit{GATsBi}. 
%Finally, cross-domain studies showed the potential of the behavioral modeling components (e.g.\ anticipation, perception decay) of \textit{GATsBi} to be applied to pedestrians as well. 
%A limitation here is using the same physics-based models for bicycle motion in the use case of pedestrians. However, it should be noted that \textit{const\_v} and \textit{const\_a} models are still pedestrian-relevant, unlike the bicycle kinematics models. 
%This should be handled by incorporating more pedestrian-relevant physics-based models in the deep ensemble of the proposed \textit{physics\_module}. 
%Future work entails further interpretability analysis for our proposed framework.

\newpage

%%%%%%%%%%%%%%%%%%%%%%%%%%%%%%%%%%%%%%%%%%%%%%%%%%%%%%%%%%%%
%%%%%%%%%%%%%%%%%%%%%%%%%%%%%%%%%%%%%%%%%%%%%%%%%%%%%%%%%%%%
%%%%%%%%%%%%%%%%%%%%%%%%%%%%%%%%%%%%%%%%%%%%%%%%%%%%%%%%%%%%

%\section*{References}

{\small
\bibliographystyle{unsrt}
\bibliography{references}
}

%%%%%%%%%%%%%%%%%%%%%%%%%%%%%%%%%%%%%%%%%%%%%%%%%%%%%%%%%%%%

\newpage
\appendix

\section{Technical Appendices and Supplementary Material}
%Technical appendices with additional results, figures, graphs and proofs may be submitted with the paper submission before the full submission deadline (see above), or as a separate PDF in the ZIP file below before the supplementary material deadline. There is no page limit for the technical appendices.

\subsection{Mass-Cycling Experiment}
We conducted a mass-cycling experiment during a conference workshop at our university, and video-captured the experiment with a drone aerially. The volunteering participants were all informed that they would be recorded and gave their written consent. For the experiment, a circular track was chosen for the following reasons. (a) All bicycles can be observed via a single drone. (b) The homogeneity of the road isolates the inter-bicycles interactions as the sole factor of their observed driving behavior. (c) The closed-loop nature of the experiment allows to control the traffic density on the road by adding or removing bicycles to simulate different situations and disruptions/disturbances.

In total, 9 video files (covering 30 minutes) were recorded at a resolution of 3840x2160 pixels, and a frame rate of 25 fps. The videos were stored in MP4 format and are about 25.7 GB large. Due to interruptions by trucks, cars, and drone landing for battery change, only some parts of the video are useful for the purpose of this investigation. Therefore, specific sequences of these videos were manually selected for dataset generation.

\begin{figure} [!ht]
    \centering
    \includegraphics[width=1.0\linewidth]{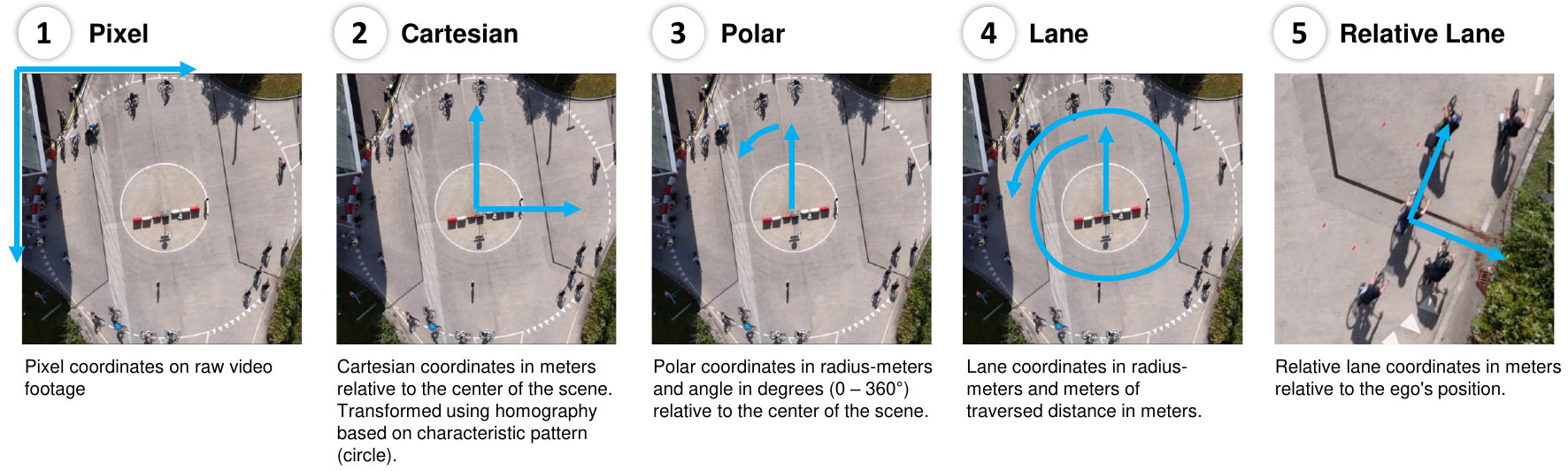}
    \caption{\textbf{Coordinate Transforms.} The coordinates from bicycles have been derived using object detection algorithms in pixel coordinates (1) and then converted to other coordinate systems (2-4) until reaching the relative lane coordinates (5) used for training the trajectory prediction model.}
    \label{fig:appendix_coordinate_transform}
\end{figure}

Two different computer vision approaches and manual annotation to detect bicycles on the aerial images: (a) object detection with YOLO, (b) an approach that compares two consecutive frames for differences to identify moving objects with OpenCV. Also, we extracted a characteristic pattern (the inner circle) with known geometric properties (radius 5.0m) using Hough transform, in order to conduct a homography transformation from pixel to Cartesian coordinates. The different coordinate systems used are illustrated in Figure~\ref{fig:appendix_coordinate_transform}.
Afterwards, we used a computational pipeline to extract trajectories from these object detections. The trajectories were filtered with a Kalman-Filter and checked for quality manually.
The Cartesian coordinates are then transformed to polar coordinates, as shown in Figure~\ref{fig:appendix_coordinate_transform}. The polar $x$ and $y$ axes represent the angle and radius (i.e.\ distance to the center of the circular track) respectively. Then, lane coordinates are computed; where $x$ and $y$ axes denote the radius and the track-aligned distance covered by the bicycle. The relative lane coordinates are the inputs to the proposed \textit{GATsBi} framework. These are computed relative to the perspective of a selected ego bicycle. 

\subsection{Social Feature Generation}

In the current implementation, the ego's neighborhood consists of at most 5 neighboring bicycles in a distance of less then 20m.
The relatively large threshold of 20m was chosen to cover longer prediction horizons up to 4s and mediocre bicycle velocities.
The adjacency matrix comprises four features for each edge, including the Cartesian distance, relative orientation, and relative speeds in horizontal and vertical directions, all relative to the ego.

\subsection{Multimodal Trajectory Sampling}
\begin{figure} [!ht]
    \centering
    \includegraphics[width=1.0\linewidth]{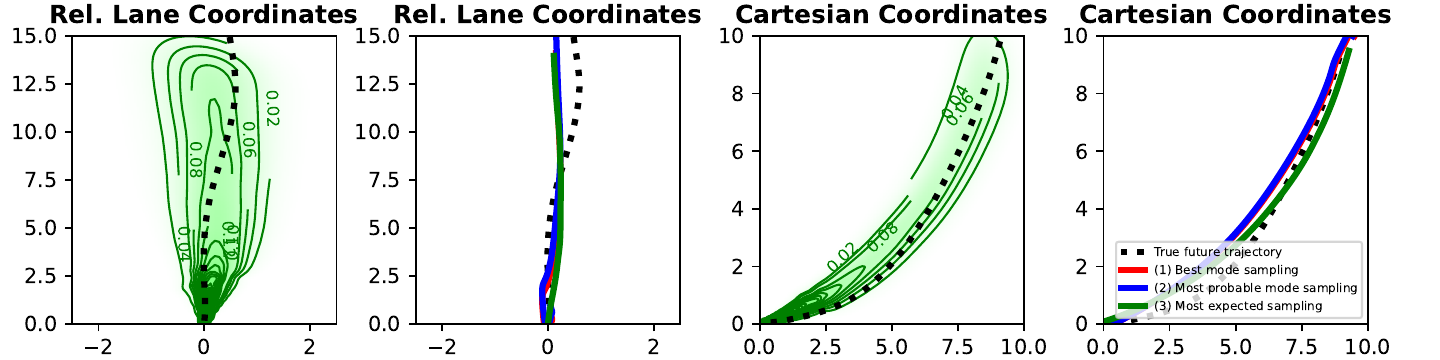}
    \caption{\textbf{Sampling Techniques for Gaussian Mixture Models.} The result of multimodal trajectory forecasting is a probability distribution. To derive a final prediction, three different sampling techniques can be used.}
    \label{fig:appendix_sampling}
\end{figure}

Since \textit{GATsBi}'s multimodal output module is a Gaussian mixture model, it presents various techniques to sample the predicted trajectory distribution $P(\hat{X}_i^p)$. Figure~\ref{fig:appendix_sampling} shows the three proposed sampling techniques.
\begingroup
\renewcommand\labelenumi{(\theenumi)}
\begin{enumerate}
    \item \textbf{Best mode sampling: } We choose the Gaussian mixture component that is closest to the ground truth (i.e.\ least ADE). The limitation here is the lack of ground truth during inference time.
    \item \textbf{Most probable mode sampling: } We choose the Gaussian mixture component with the highest mixture weight at the last prediction timestep. %This sampling technique would be biased towards the FDE metric.
    \item \textbf{Most expected sampling: } This technique is used in all our previously presented results, as expressed in \eqref{eq:expected_sampling}. This samples the expected trajectory out of the probability distribution $P(\hat{X}_i^p)$; where a linear combination of all Gaussian mixture components is computed. %This sampling technique would be biased towards the ADE metric.
\end{enumerate}
\endgroup

\begin{table}[!ht]
    \caption{\textbf{Great GATsBi Multimodal Sampling Techniques on Mass Cycling Dataset.} 
    Evaluation metrics (ADE and FDE) reported in average and standard deviation (in brackets) across all train/test splits for four different prediction horizons (1s to 4s).}
    \label{tab:app_sampling_tech}
    \centering
    \begin{tabularx}{\textwidth}{
        l
        c
        >{\centering\arraybackslash}X
        >{\centering\arraybackslash}X
        >{\centering\arraybackslash}X
        >{\centering\arraybackslash}X
        c
        >{\centering\arraybackslash}X
        >{\centering\arraybackslash}X
        >{\centering\arraybackslash}X
        >{\centering\arraybackslash}X
    }
        \toprule
        \textbf{Method} & & \multicolumn{4}{c}{\textbf{ADE}} & & \multicolumn{4}{c}{\textbf{FDE}} \\
        & & 1s & 2s & 3s & 4s & & 1s & 2s & 3s & 4s \\
        \cmidrule(lr){3-6} \cmidrule(lr){8-11}
        \textbf{unimodal GATsBi} & & 0.0757 \newline [0.0059] & 0.2180 \newline [0.0125] & 0.4302 \newline [0.0343] & 0.7760 \newline [0.0945] & & 0.1948 \newline [0.0138] & 0.5051 \newline [0.0253] & 1.2286 \newline [0.1554] & 2.3142 \newline [0.4071] \\	
        \textbf{multimodal GATsBi} & & & & & & & & & & \\
        \;\; most expected & & 0.0715 \newline [0.0066] & 0.2078 \newline [0.0130] & 0.4181 \newline [0.0354] & 0.7543 \newline [0.0960] & & 0.1893 \newline [0.0153] & 0.4891 \newline [0.0258] & 1.2641 \newline [0.1762] & 2.3827 \newline [0.4103] \\
        \;\; most probable mode & & 0.0741 \newline [0.0069] & 0.4010 \newline [0.2143] & 1.2569 \newline [0.4460] & 1.5485 \newline [0.4406] & & 0.1905 \newline [0.0159] & 0.4892 \newline [0.0282] & 1.2165 \newline [0.1670] & 2.2888 \newline [0.4390] \\
        \;\; best mode & & 0.0587 \newline [0.0047] & 0.2214 \newline [0.0136] & 0.5574 \newline [0.0909] & 0.8734 \newline [0.2342] & & 0.1598 \newline [0.0109] & 0.5181 \newline [0.1028] & 1.7967 \newline [0.5601] & 2.3795 \newline [0.6253] \\

        \bottomrule
    \end{tabularx}
\end{table}

Table~\ref{tab:app_sampling_tech} illustrates the ADE and FDE metrics for \textit{GATsBi} over various prediction horizon lengths (1, 2, 3, and 4 seconds). As previously discussed, multimodal models capture the unpredictability of human future behavior and, therefore, achieve outperform \textit{unimodal GATsBi}. 
It is interesting to note that the FDE shown by \textit{unimodal GATsBi} for 3- and 4-second prediction horizons is lower than the full \textit{GATsBi} using most expected sampling. This makes the analysis biased towards the ADE metric. Still, \textit{unimodal GATsBi} is also trained on the ADE loss; thereby, yielding a fair comparison. 
On the other hand, \textit{GATsBi} using the most probable sampling technique yields a lower FDE than \textit{unimodal GATsBi}. This sampling technique is heavily biased towards the FDE metric since it naturally chooses the mode with the highest mixture weight at the last prediction timestep.
As observed in Figure~\ref{fig:appendix_sampling}, the most expected sampling technique has, on average, the least deviations away from ground truth in Cartesian coordinates (see rightmost plot). This is in contrast to the most probable sampling technique, which has the least deviation away from the ground truth only at the last prediction timestep.
Regarding the best mode sampling technique, it produces a significantly lower ADE and FDE metrics for 1-second prediction horizon than other sampling techniques. Yet, it performs worse for longer prediction horizons and with higher variances. This is yet another limitation for this sampling method.
Hence, the choice of the sampling method is dependent on the performance metric of choice.

\subsection{Pedestrian Experiment}

A limitation here is using the same physics-based models for bicycle motion in the use case of pedestrians. However, it should be noted that \textit{const\_v} and \textit{const\_a} models are still pedestrian-relevant, unlike the bicycle kinematics models. 
This should be handled by incorporating more pedestrian-relevant physics-based models in the deep ensemble of the proposed \textit{physics\_module}, or excluding kinematic and Kalman filtering approaches from the ensemble. 
Nonetheless, \textit{GATsBi}'s physics module was able to learn the importance of the relevant physics-based models only.

%%%%%%%%%%%%%%%%%%%%%%%%%%%%%%%%%%%%%%%%%%%%%%%%%%%%%%%%%%%%

\newpage
\section*{NeurIPS Paper Checklist}

\begin{enumerate}

\item {\bf Claims}
    \item[] Question: Do the main claims made in the abstract and introduction accurately reflect the paper's contributions and scope?
    \item[] Answer: \answerYes{} % Replace by \answerYes{}, \answerNo{}, or \answerNA{}.
    \item[] Justification: The abstract and introduction clearly state the main contributions and scope of the paper. %\justificationTODO{}

\item {\bf Limitations}
    \item[] Question: Does the paper discuss the limitations of the work performed by the authors?
    \item[] Answer: \answerYes{} % Replace by \answerYes{}, \answerNo{}, or \answerNA{}.
    \item[] Justification: The limitations are discussed in the technical appendix, and details on computational runtime expense in the implementation details. % \justificationTODO{}

\item {\bf Theory assumptions and proofs}
    \item[] Question: For each theoretical result, does the paper provide the full set of assumptions and a complete (and correct) proof?
    \item[] Answer: \answerNA{} % Replace by \answerYes{}, \answerNo{}, or \answerNA{}.
    \item[] Justification: The paper has no theoretical contributions. %\justificationTODO{}

    \item {\bf Experimental result reproducibility}
    \item[] Question: Does the paper fully disclose all the information needed to reproduce the main experimental results of the paper to the extent that it affects the main claims and/or conclusions of the paper (regardless of whether the code and data are provided or not)?
    \item[] Answer: \answerYes{} % Replace by \answerYes{}, \answerNo{}, or \answerNA{}.
    \item[] Justification: We have disclosed the details of our experiments and will provide code and data publicly in the form of an extensive Github repository upon publication for reproducibility and results verification. % \justificationTODO{}

\item {\bf Open access to data and code}
    \item[] Question: Does the paper provide open access to the data and code, with sufficient instructions to faithfully reproduce the main experimental results, as described in supplemental material?
    \item[] Answer: \answerYes{} % Replace by \answerYes{}, \answerNo{}, or \answerNA{}.
    \item[] Justification: We will publicly provide the code and data on our Github repository upon publication, along with documentation and setup instructions. %\justificationTODO{}

\item {\bf Experimental setting/details}
    \item[] Question: Does the paper specify all the training and test details (e.g., data splits, hyperparameters, how they were chosen, type of optimizer, etc.) necessary to understand the results?
    \item[] Answer: \answerYes{} % Replace by \answerYes{}, \answerNo{}, or \answerNA{}.
    \item[] Justification: The implementation details are discussed in the Experiments section and are also provided within our codes. %\justificationTODO{}

\item {\bf Experiment statistical significance}
    \item[] Question: Does the paper report error bars suitably and correctly defined or other appropriate information about the statistical significance of the experiments?
    \item[] Answer: \answerYes{} % Replace by \answerYes{}, \answerNo{}, or \answerNA{}.
    \item[] Justification: In Tables 1, 2, and 4, we have included standard deviations for 5-fold cross-validation splits. % \justificationTODO{}

\item {\bf Experiments compute resources}
    \item[] Question: For each experiment, does the paper provide sufficient information on the computer resources (type of compute workers, memory, time of execution) needed to reproduce the experiments?
    \item[] Answer: \answerYes{} % Replace by \answerYes{}, \answerNo{}, or \answerNA{}.
    \item[] Justification: The implmenetation details in the paper, and the code's documentation (README.md) state the computing resources used. %\justificationTODO{}
    
\item {\bf Code of ethics}
    \item[] Question: Does the research conducted in the paper conform, in every respect, with the NeurIPS Code of Ethics \url{https://neurips.cc/public/EthicsGuidelines}?
    \item[] Answer: \answerYes{} % Replace by \answerYes{}, \answerNo{}, or \answerNA{}.
    \item[] Justification: We have reviewed and followed the NeurIPS Code of Ethics. %\justificationTODO{}

\item {\bf Broader impacts}
    \item[] Question: Does the paper discuss both potential positive societal impacts and negative societal impacts of the work performed?
    \item[] Answer: \answerYes{} % Replace by \answerYes{}, \answerNo{}, or \answerNA{}.
    \item[] Justification: The broader impact and motivation of research is outlined in the Introduction. Improving the accuracy of bicycle trajectory prediction gives rise to protecting them as vulnerable road users; especially in autonomous driving environments. %\justificationTODO{}
    
\item {\bf Safeguards}
    \item[] Question: Does the paper describe safeguards that have been put in place for responsible release of data or models that have a high risk for misuse (e.g., pretrained language models, image generators, or scraped datasets)?
    \item[] Answer: \answerNA{} % Replace by \answerYes{}, \answerNo{}, or \answerNA{}.
    \item[] Justification: 
    Our method and related mass cycling experiment data do not have a risk of misuse.
    %To protect the privacy of the mass-cycling experiment participants, the video captures are not released for this work. %\justificationTODO{}

\item {\bf Licenses for existing assets}
    \item[] Question: Are the creators or original owners of assets (e.g., code, data, models), used in the paper, properly credited and are the license and terms of use explicitly mentioned and properly respected?
    \item[] Answer: \answerYes{} % Replace by \answerYes{}, \answerNo{}, or \answerNA{}.
    \item[] Justification: We are releasing our code and data with the proper license (MIT license). %\justificationTODO{}

\item {\bf New assets}
    \item[] Question: Are new assets introduced in the paper well documented and is the documentation provided alongside the assets?
    \item[] Answer: \answerYes{} % Replace by \answerYes{}, \answerNo{}, or \answerNA{}.
    \item[] Justification: We have properly documented our repository via Readme files; including implementation details, and setup and execution instructions. % \justificationTODO{}

\item {\bf Crowdsourcing and research with human subjects}
    \item[] Question: For crowdsourcing experiments and research with human subjects, does the paper include the full text of instructions given to participants and screenshots, if applicable, as well as details about compensation (if any)? 
    \item[] Answer: \answerNA{} % Replace by \answerYes{}, \answerNo{}, or \answerNA{}.
    \item[] Justification: The contributions of the paper do not relate to human subjects. Still, the participants of the mass-cycling experiments provided written consent. %\justificationTODO{}

\item {\bf Institutional review board (IRB) approvals or equivalent for research with human subjects}
    \item[] Question: Does the paper describe potential risks incurred by study participants, whether such risks were disclosed to the subjects, and whether Institutional Review Board (IRB) approvals (or an equivalent approval/review based on the requirements of your country or institution) were obtained?
    \item[] Answer: \answerNA{} % Replace by \answerYes{}, \answerNo{}, or \answerNA{}.
    \item[] Justification: This does not apply to our research. %\justificationTODO{}

\item {\bf Declaration of LLM usage}
    \item[] Question: Does the paper describe the usage of LLMs if it is an important, original, or non-standard component of the core methods in this research? Note that if the LLM is used only for writing, editing, or formatting purposes and does not impact the core methodology, scientific rigorousness, or originality of the research, declaration is not required.
    %this research? 
    \item[] Answer: \answerNA{} % Replace by \answerYes{}, \answerNo{}, or \answerNA{}.
    \item[] Justification: This does not relate to our research and contributions. %\justificationTODO{}

\end{enumerate}

\end{document}